%% file: main.tex
\documentclass[conference]{IEEEtran}
\IEEEoverridecommandlockouts
\usepackage{cite}
\usepackage{amsmath,amssymb,amsfonts}
\usepackage{algorithmic}
\usepackage{graphicx}
\usepackage{textcomp}
\usepackage{xcolor}

\def\BibTeX{{\rm B\kern-.05em{\sc i\kern-.025em b}\kern-.08em
    T\kern-.1667em\lower.7ex\hbox{E}\kern-.125emX}}
\begin{document}

\title{Accelerating Biological Spatial Cluster Analysis with the Parallel Integral Image Technique
}

\author{\IEEEauthorblockN{1\textsuperscript{st} Seth Ockerman}
\IEEEauthorblockA{\textit{School of Computing} \\
\textit{Grand Valley State University}\\
Grand Rapids, Michigan, United States \\
ockermas@mail.gvus.edu}

\and

\IEEEauthorblockN{2\textsuperscript{nd} Zachary Klamer}
\IEEEauthorblockA{Department of Cell Biology \\
\textit{Van Andel Institute}\\
Grand Rapids, Michigan, United States \\
zachary.klamer@vai.org}
\and

\IEEEauthorblockN{2\textsuperscript{nd} Brian Haab}
\IEEEauthorblockA{\textit{Department of Cell Biology} \\
\textit{Van Andel Institute}\\
Grand Rapids, Michigan, United States \\
brian.haab@vai.org}

}

\maketitle
\input{text-sections/abstract}

\input{text-sections/introduction}
\input{text-sections/background}
\input{text-sections/cost-analysis}
\input{text-sections/results}
\input{text-sections/conclusion}

\bibliographystyle{IEEEtran}
\bibliography{main}
\end{document}

%% file: text-sections/abstract.tex
\begin{abstract}
Spatial cluster analysis (SCA) offers valuable insights into biological images; a common SCA technique is sliding window analysis (SWA). Unfortunately, SWA's computational cost hinders its application to larger images, limiting its use to small-scale images. With advancements in high-resolution microscopy, images now exceed the capabilities of previous SWA approaches, reaching sizes up to 70,000 by 85,000 pixels. To overcome these limitations, this paper introduces the parallel integral image approach to SWA, surpassing previous methods. We achieve a remarkable speedup of 131,806x on small-scale images and consistent speedups of over 10,000x on a variety of large-scale microscopy images. We analyze the computational complexity advantages of the parallel integral image approach and present experimental results that validate the superior performance of integral-image-based methods. Our approach is made available as an open-source Python PIP package available at https://github.com/OckermanSethGVSU/BioPII. 

\vspace{3pt}

Index Terms -- Sliding Window Analysis, HPC, Integral Image, Cost Analysis, Bioinformatics, Microscopy Image Analysis 
\end{abstract}

%% file: text-sections/introduction.tex
\section{Introduction}
Spatial cluster analysis (SCA) is a useful tool for studying the distribution of biological entities. One approach to SCA involves using a sliding window to identify significant cell clusters. Sliding window image analysis (SWA) divides an image into overlapping regions and applies an analysis algorithm to each region. SWA has been used in various biological applications, including predicting pancreatic cancer outcomes \cite{wisniewski_2023}, automating cell segmentation \cite{xing_2016}, and identifying candidates for medical interventions \cite{nolan_2021}. However, SWA's computational cost becomes problematic for large images, and previous attempts to improve naive-SWA have limitations due to SWA's inherent complexity. This work proposes the introduction of the parallel integral image technique \cite{veksler_2003} to biology to reduce SWA's algorithmic complexity and enable constant time computations of variable window sizes. This paper presents algorithmic and practical analysis of SWA, introduces a biology-focused parallel integral image technique \cite{veksler_2003}, and provides an open-source Python Pip package called \texttt{BioPII} for seamless integration with existing bio-image analysis software. To the authors' knowledge, this work is the first to focus on algorithmic complexity analysis of SWA and the introduction of the integral image technique to bio-image analysis.

%% file: text-sections/background.tex
\section{Background}
\label{sec:background}
Before analysing naive-SWA, integral image SWA, and  parallel integral image SWA, a brief overview of each approach and its past work is needed. 
\subsection{Naive Approaches to Sliding Window Sums}
For the purpose of this paper, naive-SWA is defined as follows: a fixed-size window is defined and moved over the entire image. At each position, an analysis algorithm is applied  to the current window and the result is stored in a new array. In this work, we focus on a simple window sum algorithm because of its prevalence. However, \texttt{BioPII} supports intergral-image-based average and standard deviation algorithms as well. Regardless of the specific algorithm, naive-SWA is often prohibitively expensive. Past work has tried to address this through parellization, achieving speedups ranging from 5x \cite{Snytsar_2020} to 156x \cite{kreuter_2010}. 

\subsection{Integral Image Approaches to Sliding Window Sums}
The integral image technique \cite{veksler_2003} is a popular computer vision tool for object detection, feature extraction, and image enhancement that has not been popularized in bioinformatics. It computes a table of cumulative sums of pixel values, where each element in the table is the sum of all the pixels above and to the left of its position in the original image. The summed area table enables efficient computation of the sum of pixel values within a rectangular region of any size. This is achieved through a simple lookup at the region's corners and subtracting the values at the relevant intermediate points. 



%% file: text-sections/cost-analysis.tex
\section{Cost Analysis of SWA Approaches}
\label{sec:cost-analysis}
Section \ref{sec:cost-analysis} analyzes the algorithmic complexity of different SWA approaches and provides insight into each approaches behavior as the size of the input grows. 
 

\subsection{Naive Approachs}

Naive-SWA has a computational complexity of $O((r - w + 1) * (c - w + 1) * w^2)$, where $r$ represents the number of rows in the target image, $c$ indicates the number of columns, and $w$ is the window size. As the image and window size increase, the cost of the naive approach grows exponentially, making it impractical for large-scale images and window sizes. To address this issue, previous work has parallelized the algorithm, resulting in a theoretical computational complexity of $O((r - w + 1) * (c - w + 1) / p * w^2)$, where $p$ is the number of parallel workers. However, achieving this level of speedup in practice is limited by overhead, GPU memory limitations, and bottlenecks associated with memory transfer costs.


\subsection{Integral Image Approach}

 The computational complexity of the integral image method for SWA is $O(r * c) + 4(r * c)$. The integral image SWA approach is multiple orders of magnitude faster than any naive SWA approach and is not affected by window size. The integral image technique can be further improved through parallelization. This technique has a computational complexity of $O(r * c) / p + 4(r * c) / p$ where p represents the number of parallel workers. The parallel integral image technique approach's cost grows significantly slower as image size increases than any of the naive approaches and is unaffected by window size.


%% file: text-sections/results.tex
\section{Runtime Testing}
\label{sec:results}
\begin{figure}[h]
 \centering
  \includegraphics[scale=0.50]{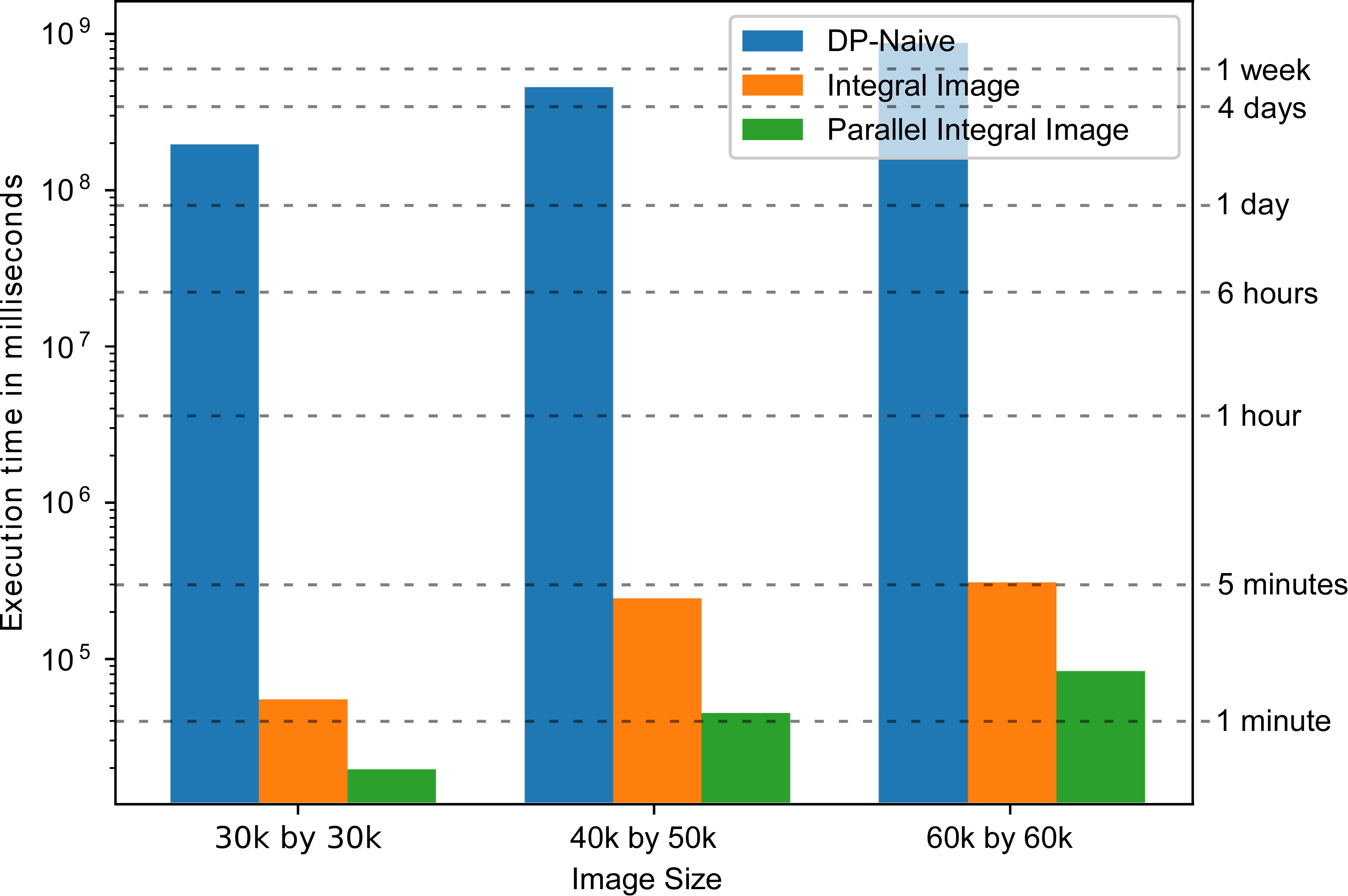}
\vspace{-0.15in}
\caption{SWA Runtime Experiment Results}
\label{fig:runtime-graph}
\end{figure}

Section \ref{sec:results} details the setup and results of runtime experiments with the different SWA approaches.  The results are summarized in table \ref{tab:runTimeTable} and shown visually in figure \ref{fig:runtime-graph}.

\subsection{Experimental Setup}
We perform SWA on immunofluorescence images of pancreatic ductal adenocarcinoma whole tumor sections to compare runtimes of naive-SWA, integral image SWA, and parallel integral image (referred to as P. integral image in table \ref{tab:runTimeTable}) SWA. The images were of sizes $30,000  \text{ x } 30,000$, $40,000  \text{ x } 50,000$, and $60,000  \text{ x } 60,000$. Our SWA experiments use window sizes which correspond to sizes of relevant cellular features: 50 (single cell), 500 (10-cell group), 1000 (gland), and 7000 (1mm biopsy). Experiments were conducted on a machine equipped with an AMD Ryzen Threadripper 1950X 16-Core Processor, 64 GB of RAM, and an NVIDIA TITAN-V graphics card.

In initial tests, we found that the naive-SWA benchmark was impractical due to its extreme runtime. On a $10,000  \text{ x } 10,000$ image, the naive approach took nearly 29 hours to complete SWA, while the parallel integral image approach finished in just 766ms (a speedup of 131,806x). Since testing the naive-SWA on large images was not feasible, we introduce DP-naive-SWA as an alternative approach that utilizes dynamic programming to significantly reduce the number of additions required. DP-naive-SWA outperformed naive-SWA by a factor of 400x on small images and will serve as the baseline approach.

    


\subsection{Results}

The integral image SWA approaches consistently outperform DP-naive-SWA on all image sizes. For images of size $30,000  \text{ x } 30,000$, $40,000  \text{ x } 50,000$ , and $60,000  \text{ x } 60,000$, the sequential integral image approach achieves speedups of 3728x, 1942x, and 2953x respectively. The parallel integral image further accelerates SWA, achieving speedups of 10,497x, 10,559x, and 10,901x on the same images previously detailed. Fundamentally, the speed of the integral image approaches enable SWA on larger scale images compared to previous methods. 

\begin{table}[h]
\centering
\begin{tabular}{|c|c|c|c|}
\hline
\textbf{Image Size} & \textbf{DP-Naive} & \textbf{Integral Image} & \textbf{P. Integral Image} \\
\hline
30k by 30k         & 211,381,433       & 56,693       & 20,137       \\
\hline
40k by 50k         & 491,374,150       & 253,082      & 46,532       \\
\hline
60k by 60k         & 943,858,845       & 319,666      & 86,583       \\
\hline
\end{tabular}
\vspace{3pt}
\caption{Runtimes of SWA on various sized images in milliseconds}
\label{tab:runTimeTable}
\end{table}



%% file: text-sections/conclusion.tex
\section{Conclusion}
\label{sec:conclusion}
We presented integral image SWA approaches that surpassed the performance of all previous methods. The parallel integral image approach significantly accelerates SWA for various biological images, thereby making a significant impact on large-scale bioanalysis. Through algorithmic analysis and runtime experimentation, we demonstrate that the integral image SWA approaches are significantly faster than the naive approach, with speedups several orders of magnitude higher than past improvements. The techniques detailed in this paper are available as the \texttt{BioPII} PIP package. Future work will explore expanding \texttt{BioPII} by supporting additional SWA algorithms and enabling calculation of multiple window sizes from a single integral image.